\documentclass{article}

\usepackage[final]{neurips_2018}

\usepackage[utf8]{inputenc} 
\usepackage[T1]{fontenc}    
\usepackage{hyperref}       
\usepackage{url}            
\usepackage{booktabs}       
\usepackage{amsfonts}       
\usepackage{nicefrac}       
\usepackage{microtype}      
\usepackage{algorithm}
\usepackage{algpseudocode}
\usepackage{graphicx}
\title{Playing 2048 With Reinforcement Learning}

\author{%
  Shilun Li \\
  Department of Mathematics \\
  Stanford University \\
  \texttt{shilun@stanford.edu} \\
  \And
  Veronica Peng \\
  Department of Computer Science \\
  Stanford University \\
  \texttt{tpeng24@stanford.edu} \\
}

\begin{document}

\maketitle

\begin{abstract}
  The game of 2048 is a highly addictive game. It is easy to learn the game, but hard to master as the created game revealed that only about 1\% games out of hundreds million ever played have been won. In this paper, we would like to explore reinforcement learning techniques to win 2048. The approaches we have took include deep Q-learning and beam search, with beam search reaching 2048 28.5\% of time.
\end{abstract}

\section{Introduction}

Recent years, reinforcement learning has been widely applied to play various strategy games involving uncertainty including but not limited to backgammon (Tesauro 1995), card games (Zha 2019), and go (Silver, 2017). As AlphaGo continuously defeated the world champion Jie Ke in the game of Go in 2016, reinforcement learning has been proved to able to achieve superhuman performance without prior human knowledge in strategy games.

Inspired by previous application of reinforcement learning, we aim to learn the best strategy to play the game 2048. 2048 is a single player name created by Gabriele Cirulli in 2014. The game is played on a $4 \times 4$ grid. Each number displayed on the screen is a power of 2, ranging from $2^1$ to $2^{15}$. A player can move all the tiles toward one direction, and two tiles will merge into their sum if they have the same value. Right after the move, a new tile will appear on the grid. The goal of this game is to get the tile with the number 2048 or higher.

\section{Literature review}

Previous papers have explored various ways to win 2048 without human knowledge. 

Amar, et al. 2017, uses policy network to learn the best strategy to play 2048 without human knowledge. The policy network is a neural network that contains two convolutional layers with ReLU as activation function. The model takes the state of the game as input, and outputs a policy vector of length 4 with each entry representing how likely choosing a policy at the current state will lead to the highest score.  To encourage exploration of different strategies, the paper also uses $\epsilon$-greedy with $\epsilon$ as a decreasing function of the number of games played. However, the model doesn't serve to win the game: it reaches 1024 most of the time, and reaches 2048 occasionally.

Szubert, et al. 2014, models the 2048 game as an MDP, and applied temporal difference learning (TDL) on the MDP to learn the best policy. The model dwarfs the human performance by reaching 2048 with a rate of 97\%. However, the model still has a hard time reaching larger tiles (never reached 32768 or higher), so Yeh, et al. 2015, improves the model by applying multi-stage temporal difference learning with 3-ply expectimax search, and the new model reaches 32768 31.75\% of the time. 

In this paper, we would like to explore and expand the tools that could be used to win 2048 using what we have learned in class.

\section{Problem description}

\subsection{State}

\begin{figure}
  \centering
  \includegraphics[scale = 0.3]{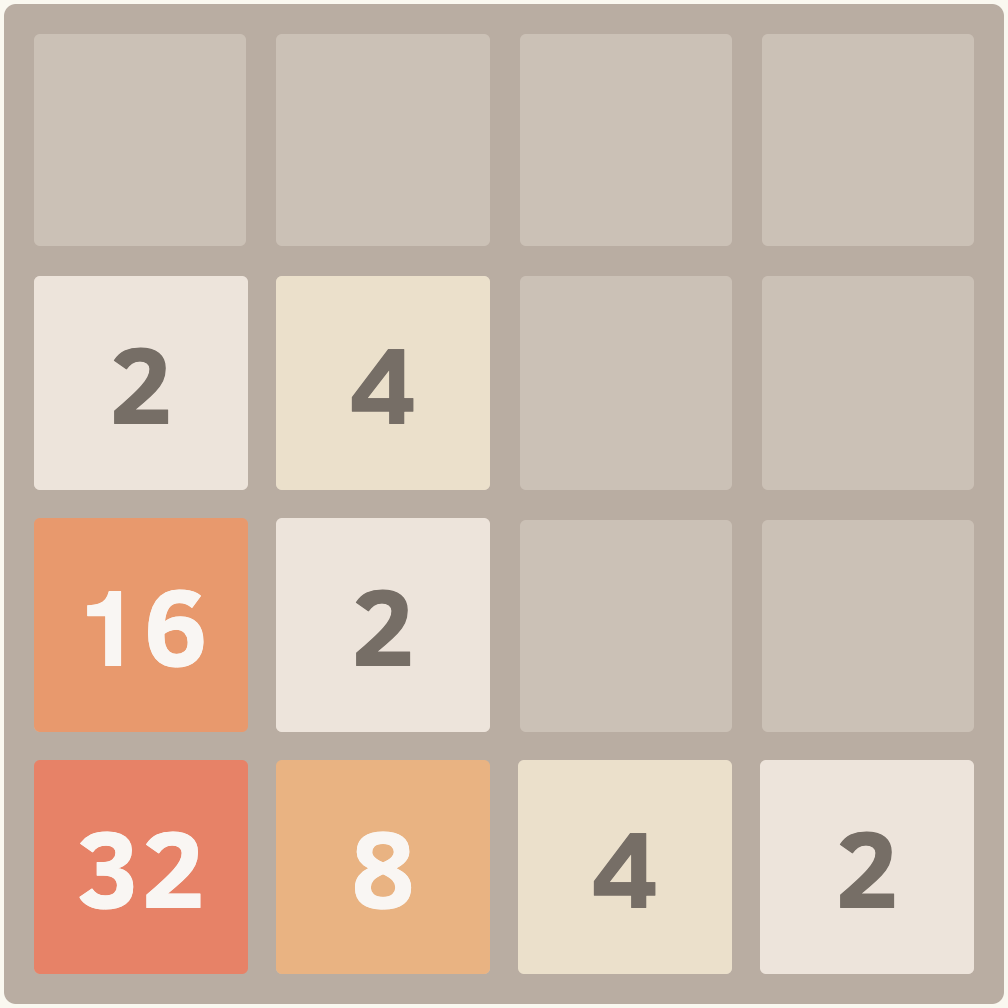}
  \caption{Illustration of the game of 2048}
\end{figure}

We use a numpy of size (16, 4, 4) to represent the state. There are thus $16 \times 4\times 4 = 256$ entries in total for the state representation. The second and third dimention ($4 \times 4$) represents each tile in the game, and the first dimension (16) represents the number on each tile. The first dimension is 16 since each tile can hold a number 0 (which means there is no number in the tile) or $2^i$ where $0 <  i < 16$. To give an example of the state representation, if the current state only has a tile 2 at position (3, 2), then the entry of numpy at (1, 3, 2) is 1, and all other entries are 0.

\subsection{Action}

There are 4 actions potentially available at each state: 1) move all the tiles to the left, 2) move all the tiles up, 3) move all the tiles to the right, and 4) move all the tiles down. However, not all the actions are available at each state. For instance, in the game in Figure 1, a player cannot move all the tiles down or to thee left. In our game, we represent all the available actions at a particular state with an list. The action list for the state in Figure 1 would be [2, 3] where 2 represent moving all the tiles up, and 3 represent moving all the tiles to the right.

\subsection{Reward}

The reward after performing an action at a state is the sum of numbers on all the merged tiles. For instance, if the we merged two tiles with number 8 and two tiles with number 2, then the reward is 20.

\subsection{Source of uncertainty}

The game 2048 is a strategy game with uncertainty. There are two sources of uncertainty, both of which are irreducible uncertainty:
\begin{itemize}
    \item The first uncertainty comes from where the new tile will appear on the grid after the move. The new tile can appear on a random empty block on the grid. The probability of it appearing on any empty block is the same.
    \item The second uncertainty comes from what number will appear on the new tile. The number on the tile is either 2 or 4. The probability for each number to appear is as follows: P(2 appears) = 0.90, P(4 appears) = 0.10.
\end{itemize}

\section{Approaches}

\subsection{Baseline}

Our baseline is to play the game by choosing a valid action at random at each state until we reach the end of the game. The strategy rarely reaches any tile above 128. (See Table 1 below for the performance of the algorithm.)

\subsection{Deep Q-learning}

We have implemented Deep Q-learning Algorithm with $\epsilon$-greedy exploration strategy to win the 2048 game. Deep Q-learning is the same as Q-learning except we use deep learning to evaluate the Q value. 

\subsubsection{Evaluating the Q value}
We tried two deep learning net to evaluate Q value, and both have similar performance. 

The first net contains 4 layers (Model 1):
\begin{enumerate}
    \item \textbf{Layer 1}: A linear layer that takes an array of size 256 as an input, and outputs an array of size $hidden$ $layer = 128$. 256 is an array flattened from the numpy representing the current state.
    \item \textbf{Layer 2}: A ReLU layer as an activation function.
    \item \textbf{Layer 3}: A dropout layer to avoid over-fitting.
    \item \textbf{Layer 4}: A linear layer that takes the output of the former layer and outputs an array of size 4. This array contains the Q-value of the 4 actions.
\end{enumerate}

The second net contains 4 layers (Model 2):
\begin{enumerate}
    \item \textbf{Layer 1}: Two convolutional layers exist in parallel in this layer. Each convolutional layer takes the state as an input and transforms it using a given kernel size. The two convolutional layers use different kernel sizes to stabilize the learning process.
    \item \textbf{Layer 2}: Four convolutional layers exist in parallel in this layer. Each convolutional layer takes the output of the former layer and transforms it using a given kernel size. The four convolutional layers use different kernel sizes to stabilize the learning process.
    \item \textbf{Layer 3}: A dropout layer to avoid over-fitting.
    \item \textbf{Layer 4}: A linear layer that takes as input the concatenation of former layer' outputs and outputs an array of size 4. This array contains the Q-value of the 4 actions.
\end{enumerate}

Each model outputs an array of size 4, representing the Q-value of the 4 actions, but not all 4 actions are available for every state. In the case where a certain action is not valid at a particular state, we remove the Q-value of the invalid action from the array before selecting the action with the highest Q value.

In order to optimize the evaluation of Q value, we minimize the Mean Squared Error Loss $$\left(r_t + \max_{a}Q(s_{t+1}, a) - Q(s_t, a_t)\right)^2$$ We keep track of the experience in a batch and back propagate when the batch size becomes 128 or the game ends. For each model, we trained for 1000 games.

\subsubsection{$\epsilon$-greedy search}
$\epsilon$-greedy search is a search algorithm that encourages exploration. $\epsilon$ is a number between 0 and 1. When selecting the best action to perform, we select a random number between 0 and 1. If the number is smaller than $\epsilon$, we choose a random action regardless of the action's Q value. Otherwise, we choose the action with the highest Q value. We used the epsilon value 0.3 in both of our models.

\subsection{Beam search}
Beam search is a greedy search algorithm that explores a tree by always choosing the $k$ best nodes. In the game of 2048, each node in the tree is a state, and each edge is an action. There are two parameters key to the beam search algorithm: 
\begin{itemize}
    \item $d$: how many levels of the tree does the algorithm want to explore. In our algorithm, we have a relatively large d ($d = 20$) since a wrong action in 2048 usually don't results immediate end of the game, but takes effect later on. We thus want the algorithm to explore the states long after the action is performed to avoid a wrong action.
    \item $k$: The algorithm keeps a list of states with the best scores at each level. $k$ is the upper limit for the number of states we can keep. In our algorithm, we have $k = 10$ to make the algorithm more likely to get out of local maximum.
\end{itemize}

At the beginning of the algorithm, it explores a list of states that can be reached using all valid actions at the current state. A random tile appears after all existing tiles have merged. As the position and the value of random tile varies, the same action at a particular state can reach scores of different states. However, evaluating all possible states takes significant computation time, and in most cases how the existing states merge after an action is much more important than the random tile in deciding whether the game ends. Thus, we randomly select one state that can be reached for each action available at the current state to construct the list of $k$ best states. We follow the same rule when we expand this list later in the algorithm.

After the initialization of the list, the algorithm repeats the following steps for each depth of the tree. We update the list from last iteration with the random states that could be reached through all valid actions for each state in the list. We then evaluate the 'goodness' of the state using a heuristic function. (A good state means it is not close to an end state of the game.) We then select the k best states from the list according to their 'goodness' score to construct the new list. After iterating for the depth $20$, we finally select the state with the highest 'goodness' score, and performs the initial action that leads to this best state.

Our heuristic function takes 4 factors into account when evaluating the the 'goodness' of the state:

\begin{table}
  \caption{Performance of all Algorithms after 1000 runs}
  \label{sample-table}
  \centering
  \begin{tabular}{lllll}
    \toprule
    Max Tile & Random Play & Deep Q-learning Model 1 & Deep Q-learning Model 2 & Beam Search \\
    \midrule
    16    & 0.3 \%  & 0 \%   & 0 \%   & 0 \%     \\
    32    & 7.1 \%  & 0 \%   & 0 \%   & 0 \%     \\
    64    & 36.9 \% & 7.4 \% & 5.7 \% & 0 \%     \\
    128   & 48.5 \% & 41.8 \%& 34.2 \%& 0 \%     \\
    256   & 7.1 \%  & 48.3 \%& 56.1 \%& 1.7 \%   \\
    512   & 0.1 \%  & 2.5 \% & 3.9 \% & 11.3 \%  \\
    1024  & 0 \%    & 0 \%   & 0.1 \% & 58.1 \%  \\
    2048  & 0 \%    & 0 \%   & 0 \%   & 28.5 \%  \\
    4096  & 0 \%    & 0 \%   & 0 \%   & 0.4 \%   \\
    \bottomrule
  \end{tabular}
\end{table}

\begin{figure}
  \centering
  \includegraphics[scale = 0.3]{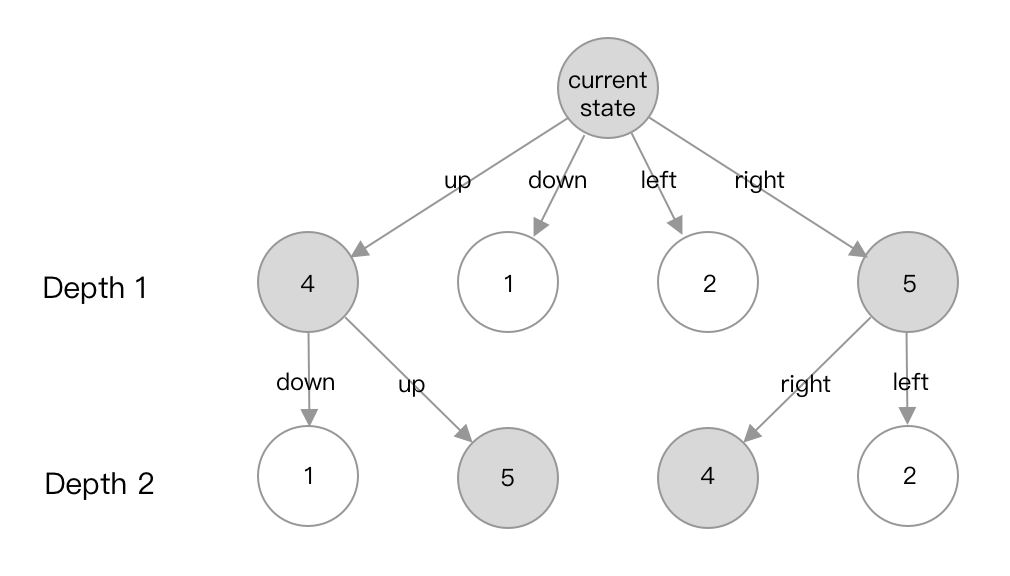}
  \caption{Illustration of the beam search algorithm. In this image, we used $k = 2$ and $d = 2$. Each edge represents an action valid at the previous stage, and each node represents a random state reached after performing the action. The number on each node is the heuristic 'goodness' score evaluated for the state. The gray nodes are the 2 nodes with the highest score stored at each depth. In this tree, we would finally perform the action 'up' since it leads to the nodes with the highest score in the last depth.}
\end{figure}

\begin{enumerate}
    \item \textbf{How many empty tiles are there in the game}: The end state of the game is when each tiles is occupied by a number. Thus, the more empty tiles there are, the better the state is.
    \item \textbf{The biggest number on any tile}: The higher the number is, the better the state is.
    \item \textbf{Smoothness}: Smoothness means whether there are any small numbers stuck in the middle of large numbers. It is hard to merge the stuck small numbers with other small numbers to get larger numbers. Thus, the smoother the state is, the better it is.
    \item \textbf{Monotonicity}: When human play the game, we usually like to stack the highest number in the corner. For the other numbers, the closer the number is to the highest number, the larger it is. In this way, it is easier for humans to merge tiles. Monotonicity measures whether the above relationship holds in the state. Intuitively, the more monotonic the state is, the better it is.
\end{enumerate}

\section{Discussion}
Note that the performance of beam search with a heuristic based on human strategy outperforms the deep Q-learning models which does not contain human knowledge. The reason behind the performance difference may be that the human strategy is very close to if not equal to the optimal strategy. We also notice that the strategy of the deep Q-learning models are approaching the human strategy as it trains. So if we can make a heuristic function of the state which incorporates how a human would evaluate the state, then a model with a simple search algorithm would have a rather good performance. On the other hand, since the deep Q-learning models does not have any prior knowledge of the game, it requires a long exploration process, therefore it may not have converged after 1000 games. 

\section{Future Work}
Since the deep Q-learning consists of 6 convolutional layers, it is rather complex and requires a decent amount of data to fully train. One of the future work for our project is training the deep Q-learning model with more data. In addition, since the prediction of the best action of a state should be consistent for any rotation of reflection of the grid. By incorporating rotations and reflections of the grid as states to train the model on, we can produce 8 times as much training data.




\section*{References}

\small
 
Amar, Jonathan, et al. "Deep Reinforcement Learning for 2048." (2017) http://www.mit.edu/~amarj/files/2048.pdf
 
Silver, David, et al. "Mastering the game of Go without human knowledge." {\it Nature550}, 354–359 (2017) doi:10.1038/nature24270

Szubert, Marcin and Wojciech Jaskowski. "Temporal Difference Learning of N-Tuple Networks for the Game 2048." {\it IEEE}, (2014) http://www.cs.put.poznan.pl/wjaskowski/pub/papers/Szubert2014\_2048.pdf

Tesauro, Gerald. "Temporal Difference Learning and TD-Gammon." {\it Communications of the ACM}, (1995) https://cling.csd.uwo.ca/cs346a/extra/tdgammon.pdf

Yeh, Kun-Hao, et al. "Multi-Stage Temporal Difference Learning for 2048-like Games." {\it TCIAIG}, (2015) https://arxiv.org/pdf/1606.07374.pdf

Zha, Daochen, et al. "RLCard: A Toolkit for Reinforcement Learning in Card Games." (2019) https://arxiv.org/abs/1910.04376
\end{document}